\documentclass[11pt]{article}					
\usepackage{colacl}					        
\usepackage{epsfig}						
\usepackage{amsmath}						

\title{Role of Verbs in Document Analysis}

\author{Judith Klavans* \and Min-Yen Kan**
\\Center for Research on Information Access* \and Department of Computer Science**
\\Columbia University
\\New York, NY  10027, USA
}

\begin{document}

\maketitle

\begin{abstract} 
We present results of two methods for assessing the event
profile of news articles as a function of verb type.  The unique
contribution of this research is the focus on the role of verbs, rather
than nouns.  Two algorithms are presented and evaluated, one of which is
shown to accurately discriminate documents by type and semantic properties,
i.e. the event profile.  The initial method, using WordNet (Miller et
al. 1990), produced multiple cross-classification of articles, primarily
due to the bushy nature of the verb tree coupled with the sense
disambiguation problem.  Our second approach using English Verb
Classes and Alternations (EVCA) Levin (1993) showed that monosemous
categorization of the frequent verbs in WSJ made it possible to usefully
discriminate documents.  For example, our results show that articles in
which communication verbs predominate tend to be opinion pieces, whereas
articles with a high percentage of agreement verbs tend to be about mergers
or legal cases.  An evaluation is performed on the results using Kendall's
\(\tau\).  We present convincing evidence for using verb semantic classes
as a discriminant in document classification.\footnote{The authors acknowledge earlier implementations by James Shaw,
and very valuable discussion from Vasileios Hatzivassiloglou, Kathleen
McKeown and Nina Wacholder.
Partial funding for this project
was provided by NSF award \#IRI-9618797
STIMULATE: Generating Coherent Summaries of On-Line Documents:
Combining Statistical and Symbolic Techniques (co-PI's McKeown and Klavans), 
and by the Columbia
University Center for Research on Information Access.}

\end{abstract} 

\section{Motivation}

We present techniques to characterize document type and event by using
semantic classification of verbs.  The intuition motivating our research is
illustrated by an examination of the role of nouns and verbs in documents.
The listing below shows the ontological categories which express the
fundamental conceptual components of propositions, using the framework of
Jackendoff (1983).  Each category permits the formation of a wh-question,
e.g. for [{\sc thing}] ``what did you buy?'' can be answered by the noun
``a fish''.  The wh-questions for [{\sc action}] and [{\sc event}] can only
be answered by verbal constructions, e.g. in the question ``what did you
do?'', where the response must be a verb, e.g. {\it jog}, {\it write}, {\it
fall}, etc.

\begin{table}[hbt]
\centering
\small
\begin{tabular}{p{2cm}p{2cm}p{2cm}} 
{\sc [thing]}		& {\sc [direction]}		& {\sc [action]} \\
{\sc [place]}		& {\sc [manner]}		& {\sc [event]} \\
{\sc [amount]} \\
\end{tabular}
\label{t:jackendoff}
\end{table}

The distinction in the ontological categories of nouns and verbs is
reflected in information extraction systems.  For example, given the noun
phrases {\it fares} and {\it US Air} that occur within a particular
article, the reader will know what the story is about, i.e. {\it fares} and
{\it US Air}.  However, the reader will not know the [{\sc event}],
i.e. what happened to the {\it fares} or to {\it US Air}.  Did airfare
prices {\it rise}, {\it fall} or {\it stabilize}?  These are the verbs most
typically applicable to prices, and which embody the event.

\subsection{Focus on the Noun}

Many natural language analysis systems focus on nouns and noun phrases in
order to identify information on who, what, and where.  For example, in
summarization, Barzilay and Elhadad (1997) and Lin and Hovy (1997) focus
on multi-word noun phrases.  For information extraction tasks, such as the
DARPA-sponsored Message Understanding Conferences (1992), only a few
projects use verb phrases (events), e.g. Appelt et al. (1993), Lin (1993).  In
contrast, the named entity task, which identifies nouns and noun phrases,
has generated numerous projects as evidenced by a host of papers in recent
conferences, (e.g. Wacholder et al. 1997, Palmer and Day 1997, Neumann et
al. 1997).  Although rich information on nominal participants, actors, and
other entities is provided, the named entity task provides no information
on {\bf what happened} in the document, i.e. the {\bf event} or {\bf
action}.  Less progress has been made on ways to utilize verbal information
efficiently.  In earlier systems with stemming, many of the verbal and
nominal forms were conflated, sometimes erroneously.  With the development
of more sophisticated tools, such as part of speech taggers, more accurate
verb phrase identification is possible.  We present in this paper an
effective way to utilize verbal information for document type
discrimination.


\subsection{Focus on the Verb}


Our initial observations suggested that both occurrence and distribution of
verbs in news articles provide meaningful insights into both article type
and content.  Exploratory analysis of parsed Wall Street Journal
data\footnote{Penn TreeBank (Marcus et al. 1994) from the Linguistic Data
Consortium.} suggested that articles characterized by movement verbs such
as {\it drop}, {\it plunge}, or {\it fall} have a different event profile
from articles with a high percentage of communication verbs, such as {\it
report}, {\it say}, {\it comment}, or {\it complain}.  However, without
associated nominal arguments, it is impossible to know whether the [{\sc
thing}] that {\it drops} refers to airfare prices or projected earnings.

In this paper, we assume that
the set of verbs in a document, when considered as a whole, can be viewed
as part of the conceptual map of the events and action in a document, in
the same way that the set of nouns has been used as a concept map for entities.  This
paper reports on two methods using verbs to determine an event profile of
the document, while also reliably categorizing documents by type.
Intuitively, the event profile refers to the classification of an article
by the kind of event.  For example, the article could be a discussion
event, a reporting event, or an argument event.

To illustrate, consider a sample article from WSJ of average length (12
sentences in length) with a high percentage of communication verbs.  The
profile of the article shows that there are 19 verbs: 11 (57\%) are
communication verbs, including {\it add}, {\it report}, {\it say}, and {\it
tell}.  Other verbs include {\it be skeptical}, {\it carry}, {\it produce},
and {\it close}.  Representative nouns include {\it Polaroid Corp.}, {\it
Michael Ellmann}, {\it Wertheim Schroder \& Co.}, {\it Prudential-Bache},
{\it savings}, {\it operating results}, {\it gain}, {\it revenue}, {\it
cuts}, {\it profit}, {\it loss}, {\it sales}, {\it analyst}, and {\it
spokesman}.

In this case, the verbs clearly contribute information that this article is
a report with more opinions than new facts.  The preponderance of
communication verbs, coupled with proper noun subjects and human nouns
(e.g.  spokesman, analyst) suggest a discussion article.  If verbs are
ignored, this fact would be overlooked.  Matches on frequent nouns like
{\it gain} and {\it loss} do not discriminate this article from one which
announces a {\it gain} or {\it loss} as breaking news; indeed, according to
our results, a breaking news article would feature a higher percentage of
motion verbs rather than verbs of communication.


\subsection{On Genre Detection}

Verbs are an important factor in providing an event
profile, which in turn might be used in categorizing articles into
different genres.  Turning to the literature in genre classification, Biber
(1989) outlines five dimensions which can be used to characterize genre.
Properties for distinguishing dimensions include verbal features such as
tense, agentless passives and infinitives.  Biber also refers to three verb
classes: private, public, and suasive verbs.  Karlgren and Cutting (1994)
take a computationally tractable set of these properties and use them to
compute a score to recognize text genre using discriminant analysis.  The
only verbal feature used in their study is present-tense verb count.  As
Karlgren and Cutting show, their techniques are effective in genre
categorization, but they do not claim to show how genres differ.  Kessler
et al. (1997) discuss some of the complexities in automatic detection of
genre using a set of computationally efficient cues, such as punctuation,
abbreviations, or presence of Latinate suffixes.  The taxonomy of genres
and facets developed in Kessler et al.  is useful for a wide range of
types, such as found in the Brown corpus.  Although some of their
discriminators could be useful for news articles (e.g. presence of second
person pronoun tends to indicate a letter to the editor), the indicators do
not appear to be directly applicable to a finer classification of news
articles.

News articles can be divided into several standard categories typically
addressed in journalism textbooks.  We base our article category ontology,
shown in lowercase, on Hill and Breen (1977), in uppercase: 
\vspace{-.2cm}

\tiny
\begin{enumerate}
\item FEATURE STORIES :   feature;
\vspace{-.2cm}
\item INTERPRETIVE STORIES:  editorial, opinion, report;
\vspace{-.2cm}
\item PROFILES;
\vspace{-.2cm}
\item PRESS RELEASES:  announcements, mergers, legal cases;
\vspace{-.2cm}
\item OBITUARIES;
\vspace{-.2cm}
\item STATISTICAL INTERPRETATION:  posted earnings;
\vspace{-.2cm}
\item ANECDOTES;
\vspace{-.2cm}
\item OTHER:  poems.
\vspace{-.2cm}
\end{enumerate}
\normalsize
\vspace{-.2cm}

The goal of our research is to identify the role of verbs, keeping in mind
that event profile is but one of many factors in determining text type. In
our study, we explored the contribution of verbs as one factor in document
type discrimination; we show how article types can be successfully
classified within the news domain using verb semantic classes. 
\vspace{-.1cm}

\section{Initial Observations}

We initially considered two specific categories of verbs in the corpus:
communication verbs and support verbs.  In the WSJ corpus, the two most
common main verbs are {\it say}, a communication verb, and {\it be}, a
support verb.  In addition to {\it say}, other high frequency communication
verbs include {\it report}, {\it announce}, and {\it state}.  In
journalistic prose, as seen by the statistics in Table~\ref{t:WSJverbs}, at
least 20\% of the sentences contain communication verbs such as {\it say}
and {\it announce}; these sentences report point of view or indicate an
attributed comment.  In these cases, the subordinated complement represents
the main event, e.g. in ``Advisors {\it announced} that IBM stock {\it
rose} 36 points over a three year period,'' there are two actions: {\it
announce} and {\it rise}.  In sentences with a communication verb as main
verb we considered both the main and the subordinate verb; this decision
augmented our verb count an additional 20\% and, even more importantly,
further captured information on the actual event in an article, not just
the communication event.  As shown in Table~\ref{t:WSJverbs}, support
verbs, such as {\it go} (``go out of business'') or {\it get} (``get
along''), constitute 30\%, and other content verbs, such as {\it fall},
{\it adapt}, {\it recognize}, or {\it vow}, make up the remaining 50\%.  If
we exclude all support type verbs, 70\% of the verbs yield information in
answering the question ``what happened?'' or ``what did X do?''

\begin{table} \small \centering
\begin{tabular}{llc} 
\hline {\bf Verb Type} & {\bf Sample Verbs} & {\bf \%} \\ 
\hline {\bf communication} & say, announce, ... & 20\% \\ 
{\bf support} & have, get, go, ... & 30\% \\ 
{\bf remainder} & abuse, claim, offer, ... & 50\% \\ 
\hline 
\end{tabular} 
\caption{Approximate Frequency of verbs by type from the {\it Wall Street
Journal} (main and selected subordinate verbs, {\it n} = 10,295).}
\label{t:WSJverbs}
\end{table}
\vspace{-.1cm}

\section{Event Profile:\hspace{-.1cm} WordNet\hspace{-.05cm} and\hspace{-.05cm} EVCA}

Since our first intuition of the data suggested that articles with a
preponderance of verbs of a certain semantic type might reveal aspects of
document type, we tested the hypothesis that verbs could be used as a
predictor in providing an event profile.  We developed two algorithms to:
(1) explore WordNet ({\tt WN-Verber}) to cluster related verbs and build a
set of verb chains in a document, much as Morris and Hirst (1991) used
Roget's Thesaurus or like Hirst and St. Onge (1998) used WordNet to build
noun chains; (2) classify verbs according to a semantic classification
system, in this case, using Levin's (1993) {\it English Verb Classes and
Alternations} ({\tt EVCA-Verber}) as a basis.  For source material, we used
the manually-parsed Linguistic Data Consortium's {\it Wall Street Journal}
(WSJ) corpus from which we extracted main and complement of communication
verbs to test the algorithms on.


\paragraph{Using WordNet.}

\begin{figure}[htb]
\centering
\epsfig{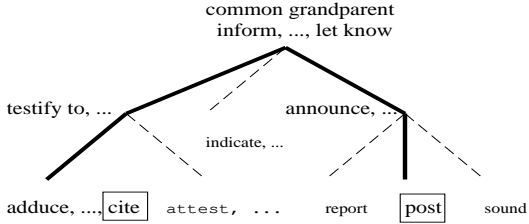}
\caption{Taxonomic Relations for {\it cite} and {\it post} in WordNet.}
\label{f:relateVerbs} 
\end{figure}

Our first technique was to use WordNet to build links between verbs and to
provide a semantic profile of the document.  WordNet is a general lexical
resource in which words are organized into synonym sets, each representing
one underlying lexical concept (Miller et al. 1990).  These synonym sets --
or synsets -- are connected by different semantic relationships such as
hypernymy (i.e. {\it plunging} is a way of {\it descending}), synonymy,
antonymy, and others (see Fellbaum 1990).  The determination of relatedness
via taxonomic relations has a rich history (see Resnik 1993 for a review).
The premise is that words with similar meanings will be located relatively
close to each other in the hierarchy. Figure~\ref{f:relateVerbs} shows the
verbs {\it cite} and {\it post}, which are related via a common ancestor
{\it inform, $\ldots$, let know}.

\paragraph{The {\tt WN-Verber} tool.}

We used the hypernym relationship in WordNet because of its high coverage.
We counted the number of edges needed to find a common ancestor for a pair
of verbs.  Given the hierarchical structure of WordNet, the lower the edge
count, in principle, the closer the verbs are semantically.  Because WordNet allows
individual words (via synsets) to be the descendent of possibly more than
one ancestor, two words can often be related by more than one common
ancestor via different paths, possibly with the same relationship
(grandparent and grandparent, or with different relations (grandparent and
uncle).  

\paragraph{Results from {\tt WN-Verber.}}

\begin{figure}[htb]
\centering
\epsfig{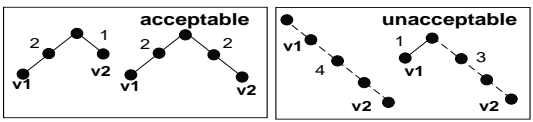}
\caption{Configurations for relating verbs in our system.}
\label{f:relateVerbs2} 
\end{figure}

We ran all articles longer than 10 sentences in the WSJ corpus (1236
articles) through {\tt WN-Verber}.  Output showed that several verbs --
e.g. {\it go}, {\it take}, and {\it say} -- participate in a very large
percentage of the high frequency synsets (approximate 30\%).  This is due
to the width of the verb forest in WordNet (see Fellbaum 1990); top level
verb synsets tend to have a large number of descendants which are arranged
in fewer generations, resulting in a flat and bushy tree structure.  For
example, a top level verb synset, {\it inform, $\ldots$, give information,
let know} has over 40 children, whereas a similar top level noun synset,
{\it entity}, only has 15 children.  As a result, using fewer than two
levels resulted in groupings that were too limited to aggregate verbs
effectively.  Thus, for our system, we allowed up to two edges to intervene
between a common ancestor synset and each of the verbs' respective synsets,
as in Figure \ref{f:relateVerbs2}. 

In addition to the problem of the flat nature of the verb hierarchy, our
results from {\tt WN-Verber} are degraded by ambiguity; similar effects
have been reported for nouns.  Verbs with differences in high versus low
frequency senses caused certain verbs to be incorrectly related; for
example, {\it have} and {\it drop} are related by the synset meaning ``to
give birth'' although this sense of {\it drop} is rare in WSJ.
 
The results of {\tt WN-Verber} in Table~\ref{t:WordNetOverview} reflect
the effects of bushiness and ambiguity.  The five most frequent synsets are
given in column 1; column 2 shows some typical verbs which participate in
the clustering; column 3 shows the type of article which tends to contain
these synsets.  Most articles (864/1236 = 70\%) end up in the top five
nodes.  This illustrates the ineffectiveness of these most frequent WordNet
synset to discriminate between article types.

\vspace{-.2cm}


\begin{center}
\begin{table}[htb]
\tiny
\begin{tabular}{p{2.2cm}p{1.6cm}p{2.8cm}}
\hline
{\bf Synset} & {\bf Sample Verbs} 

in Synset & {\bf Article types} 

(listed in order) \\
\hline
{\bf Act} 

(interact, act together, ...) \vspace{.2cm} & 
have, relate, give, tell &
announcements, editorials, features \\
{\bf Communicate} 

(communicate, intercommunicate, ...) \vspace{.2cm} & 
give, get, inform, tell &
announcements, editorials, features, poems \\
{\bf Change} 

(change) \vspace{.2cm} & 
have, modify, take &
poems, editorials, announcements, features \\
{\bf Alter}

(alter, change) \vspace{.2cm} & 
convert, make, get &
announcements, poems, editorials \\
{\bf Inform} 

(inform, round on, ...) \vspace{.2cm} & 
inform, explain, describe & 
announcements, poems, features \\
\hline
\end{tabular}
\normalsize
\caption{Frequent synsets and article types.}
\label{t:WordNetOverview}
\end{table}
\end{center}
\vspace{-.4cm}


\paragraph{Evaluation using Kendall's Tau.}

We sought
independent confirmation to assess the correlation between two variables'
rank for {\tt WN-Verber} results.  To evaluate the effects of one synset's frequency on another, we
used Kendall's tau (\(\tau\)) rank order statistic (Kendall 1970). For
example, was it the case that verbs under the synset {\it act} tend not to
occur with verbs under the synset {\it think}?  If so, do articles with
this property fit a particular profile?  In our results, we have
information about synset frequency, where each of the 1236 articles in the
corpus constitutes a sample.  Table~\ref{t:tauWordNet} shows the results of
calculating Kendall's \(\tau\) with considerations for ranking ties,
for all \(\binom{10}{2}\) = 45 pairing combinations of the top 10 most
frequently occurring synsets.  Correlations can range from $-1{.}0$ 
reflecting inverse correlation, to $+1{.}0$ showing direct correlation,
i.e. the presence of one class increases as the presence of the correlated
verb class increases.  A \(\tau\) value of 0 would show that the two
variables' values are independent of each other.

Results show a significant positive correlation between the synsets.  The
range of correlation is from ${.}850$ between the {\bf communication} verb
synset ({\it give}, {\it get}, {\it inform}, ...) and the {\bf act} verb
synset ({\it have}, {\it relate}, {\it give}, ...)  to ${.}238$ between
the {\bf think} verb synset ({\it plan}, {\it study}, {\it give}, ...)
and the {\bf change state} verb synset ({\it fall}, {\it come}, {\it
close}, ...).


These correlations show that frequent synsets do not behave independently
of each other and thus confirm that the WordNet results are not an
effective way to achieve document discrimination.  Although the WordNet
results were not discriminatory, we were still convinced that our initial
hypothesis on the role of verbs in determining event profile was worth
pursuing.  We believe that these results are
a by-product of lexical ambiguity and of the richness of the WordNet
hierarchy.  We thus decided to pursue a new approach to test our
hypothesis, one which turned out to provide us with clearer and more robust
results.

\begin{table}[htb]
\tiny
\begin{tabular}{@{\extracolsep{\fill}}l|p{.4cm}|p{.4cm}|p{.4cm}|p{.4cm}|p{.4cm}|p{.4cm}|p{.4cm}|p{.4cm}|p{.4cm}|p{.4cm}|}
\cline{2-10}
& act & com & chng & alter & infm & exps & thnk & judg & trnf \\
\hline \vline
state		& .407 & .296 & .672 & .461 & .286 & .269 & .238 & .355 & .268 \\
\cline{1-10} \vline
trnsf		& .437 & .436 &	.251 & .436 & .251 & .404 & .369 & .359 \\
\cline{1-9} \vline
judge		& .444 & .414 &	.435 & .450 & .340 & .348 & .427 \\
\cline{1-8} \vline
exprs		& .444 & .414 &	.435 & .397 & .322 & .432 \\
\cline{1-7} \vline
think		& .444 & .414 &	.435 & .397 & .398 \\
\cline{1-6} \vline
infrm		& .614 & .649 & .341 & .380 \\
\cline{1-5} \vline
alter		& .501 & .454 &	.619 \\
\cline{1-4} \vline
chnge		& .496 & .393 \\
\cline{1-3} \vline
comun		& .850 \\
\cline{1-2}
\end{tabular}
\normalsize
\caption{Kendall's \(\tau\) for frequent WordNet synsets.}
\label{t:tauWordNet}
\end{table}
\vspace{-.5cm}

\paragraph{Utilizing EVCA.}

A different approach to test the hypothesis was to use another semantic
categorization method; we chose the semantic classes of Levin's EVCA as a
basis for our next analysis.\footnote{Strictly speaking, our classification
is based on EVCA.  Although many of our classes are precisely defined in
terms of EVCA tests, we did impose some extensions.  For example,
support verbs are not an EVCA category.}  Levin's seminal work is based on
the time-honored observation that verbs which participate in similar
syntactic alternations tend to share semantic properties.  Thus, the
behavior of a verb with respect to the expression and interpretation of its
arguments can be said to be, in large part, determined by its meaning.
Levin has meticulously set out a list of syntactic tests (about 100 in
all), which predict membership in no less than 48 classes, each of which is
divided into numerous sub-classes.  The rigor and thoroughness of Levin's
study permitted us to encode our algorithm, {\tt EVCA-Verber}, on a sub-set
of the EVCA classes, ones which were frequent in our corpus.  First, we
manually categorized the 100 most frequent verbs, as well as 50 additional
verbs, which covers 56\% of the verbs by token in the corpus.  We subjected
each verb to a set of strict linguistic tests, as shown in
Table~\ref{t:EVCATests} and verified primary verb usage against the corpus.
\vspace{-.2cm}

\begin{center}
\begin{table}[htb]
\tiny
\begin{tabular}{p{2.2cm}p{5.3cm}}
\hline
{\bf Verb Class}

(sample verbs) & {\bf Sample Test} \\
\hline
{\bf Communication} 

(add, say, announce, ...) \vspace{.2cm} &
(1) Does this involve a transfer of ideas?

(2) X {\tt verb}ed ``something.'' \\
 
{\bf Motion} 

(rise, fall, decline, ...) \vspace{.2cm} &
(1) *``X {\tt verb}ed without moving". \\
{\bf Agreement} 

(agree, accept, concur, ...) \vspace{.2cm} &
(1) ``They {\tt verb}ed to join forces." 

(2) involves more than one participant. \\
{\bf Argument} 

(argue, debate, , ...) \vspace{.2cm} &
(1) ``They {\tt verb}ed (over) the issue.''

(2) indicates conflicting views. 

(3) involves more than one participant. \\
{\bf Causative}

(cause) \vspace{.2cm} &
(1) X {\tt verb}ed Y (to happen/happened).

(2) X brings about a change in Y. \\
\hline
\end{tabular}
\normalsize
\caption{EVCA verb class test}
\label{t:EVCATests}
\end{table}
\end{center}
\vspace{-1cm}

\paragraph{Results from {\tt EVCA-Verber.}}

In order to be able to compare article types and emphasize their
differences, we selected articles that had the highest percentage of a
particular verb class from each of the ten verb classes; we chose five
articles from each EVCA class, yielding a total of 50 articles for analysis
from the full set of 1236 articles.  We observed that each class
discriminated between different article types as shown in
Table~\ref{t:EVCAOverview}.  In contrast to Table~\ref{t:WordNetOverview},
the article types are well discriminated by verb class.  For example, a
concentration of {\bf communication} class verbs ({\it say}, {\it report},
{\it announce}, $\ldots$) indicated that the article type was a general
announcement of short or medium length, or a longer feature article with
many opinions in the text.  Articles high in {\bf motion} verbs were
also announcements, but differed from the communication ones, in that they
were commonly postings of company earnings reaching a new high or dropping
from last quarter.  {\bf Agreement} and {\bf argument} verbs appeared in
many of the same articles, involving issues of some controversy.
However, we noted that articles with agreement verbs were a superset of the
argument ones in that, in our corpus, argument verbs did not appear in
articles concerning joint ventures and mergers.  Articles marked by {\bf
causative} class verbs tended to be a bit longer, possibly reflecting prose
on both the cause and effect of a particular action.  We also used {\tt
EVCA-Verber} to investigate articles marked by the absence of members of
each verb class, such as articles lacking any verbs in the motion verb
class.  However, we found that absence of a verb class was not
discriminatory.


\begin{center}
\begin{table}[htb]
\tiny
\begin{tabular}{p{2.9cm}p{4cm}}
\hline
{\bf Verb Class}

(sample verbs) & {\bf Article types}

(listed by frequency) \\
\hline
{\bf Communication} 

(add, say, announce, ...) \vspace{.2cm} &
issues, reports, opinions, editorials \\
{\bf Motion} 

(rise, fall, decline, ...) \vspace{.2cm} &
posted earnings, announcements \\
{\bf Agreement} 

(agree, accept, concur, ...) \vspace{.2cm} &
mergers, legal cases, transactions (without buying and selling) \\
{\bf Argument} 

(argue, indicate, contend, ...) \vspace{.2cm} &
legal cases, opinions \\
{\bf Causative}

(cause) \vspace{.2cm} &
opinions, feature, editorials \\
\hline
\end{tabular}
\normalsize
\caption{EVCA-based verb class results.}
\label{t:EVCAOverview}
\end{table}
\end{center}
\vspace{-1cm}

\paragraph{Evaluation of EVCA verb classes.}


To strengthen the observations that articles dominated by verbs of one
class reflect distinct article types, we verified that the verb classes
behaved independently of each other.  Correlations for EVCA classes are
shown in Table~\ref{t:tauEVCA}.  These show a markedly lower level of
correlation between verb classes than the results for WordNet synsets, the
range being from ${.}265$ between motion and aspectual verbs to $-{.}026$
for motion verbs and agreement verbs.  These low values of \(\tau\) for
pairs of verb classes reflects the independence of the classes.  For
example, the {\bf communication} and {\bf experience} verb classes are
weakly correlated; this, we surmise, may be due to the different ways
opinions can be expressed, i.e. as factual quotes using {\bf communication}
class verbs or as beliefs using {\bf experience} class verbs.

\begin{table}[htb]
\tiny
\begin{tabular}{@{\extracolsep{\fill}}l|p{.6cm}|p{.6cm}|p{.6cm}|p{.6cm}|p{.6cm}|p{.6cm}|p{.6cm}|p{.6cm}|}
\cline{2-8}
& comun & motion & agree & argue & exp & aspect & cause \\
\hline \vline
appear		& .122 & .076 &	.077 & .072 & .182 & .112 & .037 \\
\cline{1-8} \vline
cause 		& .093 & .083 &	.000 & .000 & .073 & .096 \\
\cline{1-7} \vline
aspect		& .246 & .265 &	.034 & .110 & .189 \\
\cline{1-6} \vline
exp   		& .260 & .130 & .054 & .054 \\
\cline{1-5} \vline
argue		& .162 & .045 &	.033 \\
\cline{1-4} \vline
argree		& .071 & -.026 \\
\cline{1-3} \vline
motion		& .259 \\
\cline{1-2}
\end{tabular}
\normalsize
\caption{Kendall's \(\tau\) for EVCA based verb classes.}
\label{t:tauEVCA}
\end{table}

\vspace{-.3cm}
\section{Results and Future Work.}

\paragraph{Basis for WordNet and EVCA comparison.}  This paper reports
results from two approaches, one using WordNet and other based on EVCA
classes.  However, the basis for comparison must be made explicit.  In the
case of WordNet, all verb tokens ({\it n} = 10K) were considered in all
senses, whereas in the case of EVCA, a subset of less ambiguous verbs were
manually selected.  As reported above, we covered 56\% of the verbs by
token.  Indeed, when we attempted to add more verbs to EVCA categories, at
the 59\% mark we reached a point of difficulty in adding new verbs due to
ambiguity, e.g. verbs such as {\it get}.  Thus, although
our results using EVCA are revealing in important ways, it must be
emphasized that the comparison has some imbalance which puts WordNet in an
unnaturally negative light.  In order to accurately compare the two
approaches, we would need to process either the same less ambiguous verb
subset with WordNet, or the full set of all verbs in all senses with EVCA.
Although the results reported in this paper permitted the validation of our
hypothesis, unless a fair comparison between resources is performed,
conclusions about WordNet as a resource versus EVCA class distinctions
should not be inferred.

\paragraph{Verb Patterns.}
In addition to considering verb type frequencies in texts, we have observed
that verb distribution and patterns might also reveal subtle information in
text.  Verb class distribution within the document and within particular
sub-sections also carry meaning.  For example, we have observed that when
sentences with movement verbs such as {\it rise} or {\it fall} are followed
by sentences with {\it cause} and then a telic aspectual verb such as {\it
reach}, this indicates that a value rose to a certain point due to the
actions of some entity.  Identification of such sequences will enable us to
assign functions to particular sections of contiguous text in an article,
in much the same way that text segmentation program seeks identify topics
from distributional vocabulary \cite{Hearst94,Kanetal98}.  We can also use specific
sequences of verbs to help in determining methods for performing semantic
aggregation of individual clauses in text generation for summarization.
 
\paragraph{Future Work.}
Our plans are to extend the current research in terms of verb coverage and
in terms of article coverage.  For verbs, we plan to (1) increase the verbs
that we cover to include phrasal verbs; (2) increase coverage of verbs by
categorizing additional high frequency verbs into EVCA classes; (3) examine
the effects of increased coverage on determining article type.  For
articles, we plan to explore a general parser so we can test our hypothesis
on additional texts and examine how our conclusions scale up.  Finally, we
would like to combine our techniques with other indicators to form a more
robust system, such as that envisioned in Biber (1989) or suggested in
Kessler et al. (1997).  
\vspace{.1cm}

\paragraph{Conclusion.}
We have outlined a novel approach to document analysis for news articles
which permits discrimination of the event profile of news articles.  The
goal of this research is to determine the role of verbs in document
analysis, keeping in mind that event profile is one of many factors in
determining text type.  Our results show that Levin's EVCA verb classes
provide reliable indicators of article type within the news domain.  We
have applied the algorithm to WSJ data and have discriminated articles with
five EVCA semantic classes into categories such as features, opinions,
and announcements.  This approach to document type classification
using verbs has not been explored previously in the literature.  Our
results on verb analysis coupled with what is already known about NP
identification convinces us that future combinations of information will be
even more successful in categorization of documents.  Results such as these
are useful in applications such as passage retrieval, summarization, and
information extraction.

\footnotesize
\bibliographystyle{acl}
\bibliography{verbs}

\begin{thebibliography}{}

\bibitem[\protect\citename{Appelt \bgroup et al.\egroup }1993]{fastus}
D.~Appelt, J.~Hobbs, J.~Bear, D.~Isreal, and M.~Tyson.
\newblock 1993.
\newblock Fastus: A finite state processor for information extraction from real
  world text.
\newblock In {\em Proceedings of the 13th International Joint Conference on
  Artificial Intelligence (IJCAI)}, Chambery, France.

\bibitem[\protect\citename{Barzilay and Elhadad}1997]{Barzilay&Elhadad97}
Regina Barzilay and Michael Elhadad.
\newblock 1997.
\newblock Using lexical chains for text summarization.
\newblock In {\em Proceedings of the Intelligent Scalable Text Summarization
  Workshop (ISTS'97), {ACL}}, Madrid, Spain.

\bibitem[\protect\citename{Biber}1989]{Biber89}
Douglas Biber.
\newblock 1989.
\newblock A typology of english texts.
\newblock {\em Language}, 27:3--43.

\bibitem[\protect\citename{Fellbaum}1990]{Fellbaum90}
Christiane Fellbaum.
\newblock 1990.
\newblock English verbs as a semantic net.
\newblock {\em International Journal of Lexicography}, 3(4):278--301.

\bibitem[\protect\citename{Hearst}1994]{Hearst94}
Maarti~A. Hearst.
\newblock 1994.
\newblock Multi-paragraph segmentation of expository text.
\newblock In {\em Proceedings of the 32th Annual Meeting of the Association of
  Computational Linguistics}.

\bibitem[\protect\citename{Hill and Breen}1977]{HillandBrean77}
Evan Hill and John~J. Breen.
\newblock 1977.
\newblock {\em Reporting \& Writing the News}.
\newblock Little, Brown and Company, Boston, Massachusetts.

\bibitem[\protect\citename{Hirst and St-Onge}1998]{Hirst&StOnge98}
Graeme Hirst and David St-Onge.
\newblock 1998.
\newblock Lexical chains as representations of context for the detection and
  correction of malapropisms.
\newblock {\em WordNet: An electronic lexical database and some of its
  applications}.

\bibitem[\protect\citename{Jackendoff}1983]{Jackendoff86}
Ray Jackendoff.
\newblock 1983.
\newblock {\em Semantics and Cognition}.
\newblock MIT University Press, Cambridge, Massachusetts.

\bibitem[\protect\citename{Kan \bgroup et al.\egroup }1998]{Kanetal98}
Min-Yen Kan, Judith~L. Klavans, and Kathleen~R. McKeown.
\newblock 1998.
\newblock Linear segmentation and segment relevance.
\newblock Unpublished Manuscript.

\bibitem[\protect\citename{Karlgren and Cutting}1994]{Karlgren&Cutting94}
Jussi Karlgren and Douglass Cutting.
\newblock 1994.
\newblock Recognizing text genres with simple metrics using discriminant
  analysis.
\newblock In {\em Fifteenth International Conference on Computational
  Linguistics (COLING '94)}, Kyoto, Japan.

\bibitem[\protect\citename{Kendall}1970]{Kendall70}
Maurice~G. Kendall.
\newblock 1970.
\newblock {\em Rank Correlation Methods}.
\newblock Griffin, London, England, 4th edition.

\bibitem[\protect\citename{Kessler \bgroup et al.\egroup }1997]{Kessleretal97}
Brent Kessler, Geoffrey Nunberg, and Hinrich Sch{\"u}tze.
\newblock 1997.
\newblock Automatic detection of text genre.
\newblock In {\em Proceedings of the 35th Annual Meeting of the Association of
  Computational Linguistics}, Madrid, Spain.

\bibitem[\protect\citename{Levin}1993]{Levin93}
Beth Levin.
\newblock 1993.
\newblock {\em English Verb Classes and Alternations}.
\newblock University of Chicago Press, Chicago, Ohio.

\bibitem[\protect\citename{Lin and Hovy}1997]{Lin&Hovy97}
Chin-Yew Lin and Eduard Hovy.
\newblock 1997.
\newblock Identifying topics by position.
\newblock In {\em Proceedings of the 5th {ACL} Conference on Applied Natural
  Language Processing}, pages 283--290, Washington, D.C., April.

\bibitem[\protect\citename{Lin}1993]{circus}
Dekang Lin.
\newblock 1993.
\newblock \protect{University of Manitoba: Description of the NUBA System as
  Used for MUC-5}.
\newblock In {\em Proceedings of the Fifth Conference on Message Understanding
  MUC-5}, pages 263--275, Baltimore, Maryland. ARPA.

\bibitem[\protect\citename{Marcus~et al.}1994]{Marcusetal94}
Mitch Marcus~et al.
\newblock 1994.
\newblock {\em The Penn Treebank: Annotating Predicate Argument Structure}.
\newblock ARPA Human Language Technology Workshop.

\bibitem[\protect\citename{Miller \bgroup et al.\egroup }1990]{Milleretal90}
George~A. Miller, Richard Beckwith, Christiane Fellbaum, Derek Gross, and
  Katherine~J. Miller.
\newblock 1990.
\newblock Introduction to {WordNet}: An on-line lexical database.
\newblock {\em International Journal of Lexicography (special issue)},
  3(4):235--312.

\bibitem[\protect\citename{Morris and Hirst}1991]{Morris&Hirst91}
Jane Morris and Graeme Hirst.
\newblock 1991.
\newblock Lexical coherence computed by thesaural relations as an indicator of
  the structure of text.
\newblock {\em Computational Linguistics}, 17(1):21--42.

\bibitem[\protect\citename{MUC}1992]{muc}
1992.
\newblock {\em Message Understanding Conference --- MUC}.

\bibitem[\protect\citename{Neumann \bgroup et al.\egroup }1997]{newref97}
G{\"u}nter Neumann, Rolf Backofen, Judith Baur, Marcus Becker, and Christian
  Braun.
\newblock 1997.
\newblock An information extraction core system for real world german text
  processing.
\newblock In {\em Proceedings of the 5th {ACL} Conference on Applied Natural
  Language Processing}, pages 209--216, Washington, D.C., April.

\bibitem[\protect\citename{Palmer and Day}1997]{Palmer&Day97}
David~D. Palmer and David~S. Day.
\newblock 1997.
\newblock A statistical profile of the named entity task.
\newblock In {\em Proceedings of the 5th {ACL} Conference on Applied Natural
  Language Processing}, pages 190--193, Washington, D.C., April.

\bibitem[\protect\citename{Resnik}1993]{Resnik93}
Philip Resnik.
\newblock 1993.
\newblock {\em Selection and Information: A Class-Based Approach to Lexical
  Relationships}.
\newblock {Ph.D.} thesis, Department of Computer and Information Science,
  University of Pennsylvania.

\bibitem[\protect\citename{Wacholder \bgroup et al.\egroup
  }1997]{Wacholderetal97}
Nina Wacholder, Yael Ravin, and Misook Choi.
\newblock 1997.
\newblock Disambiguation of proper names in text.
\newblock In {\em Proceedings of the 5th {ACL} Conference on Applied Natural
  Language Processing}, volume~1, pages 202--209, Washington, D.C., April.

\end{thebibliography}

\clearpage

\cite{muc}
\cite{Barzilay&Elhadad97}
\cite{Biber89}
\cite{Jackendoff86}
\cite{Karlgren&Cutting94}
\cite{Kessleretal97}
\cite{Kendall70}
\cite{Levin93}
\cite{Marcusetal94}
\cite{Milleretal90}
\cite{Palmer&Day97}
\cite{Resnik93}
\cite{Lin&Hovy97}
\cite{Kanetal98}
\cite{Hearst94}
\cite{fastus}
\cite{Wacholderetal97}
\cite{circus}
\cite{newref97}
\cite{Hirst&StOnge98}
\cite{Morris&Hirst91}
\cite{Fellbaum90}
\cite{HillandBrean77}

\end{document}